\documentclass[
hf,
]{ceurart}

\usepackage{listings}
\usepackage{comment}
\usepackage[scientific-notation=true]{siunitx}
\usepackage{graphicx} 
\usepackage{amsmath}
\usepackage{amssymb}
\begin{document}%https://www.overleaf.com/project/62c73a8000c5c987d1682818
	
	\copyrightyear{2022}
	\copyrightclause{Copyright for this paper by its authors.
		Use permitted under Creative Commons License Attribution 4.0
		International (CC BY 4.0).}

	\setlength{\floatsep}{1.0pt} %: space left between floats (12.0pt plus 2.0pt minus 2.0pt).
	\setlength{\textfloatsep}{12.0pt plus 2.0pt minus 2.0pt} %: space between last top float or first bottom float and the text (14.0pt plus 2.0pt minus 2.0pt).
	\setlength{\intextsep}{2pt} % : space left on top and bottom of an in-text float (16.0pt plus 2.0pt minus 2.0pt).
	\setlength{\abovecaptionskip}{10pt plus 1pt minus 1pt} %space above caption 
	\setlength{\belowcaptionskip}{10pt plus 2pt minus 2pt} %space below caption ().
	
	\conference{ICCBR XCBR'22: 4th Workshop on XCBR: Case-based Reasoning for the Explanation of Intelligent Systems at ICCBR-2022, September, 2022, Nancy, France}
	
	\title{When a CBR in Hand is Better than Twins in the Bush}

	\author[1]{Mobyen Uddin Ahmed}[%
	orcid=0000-0003-1953-6086,
	email=mobyen.uddin.ahmed@mdu.se,
	url=http://www.es.mdh.se/staff/149-Mobyen\_Uddin\_Ahmed
	]
	\author[1]{Shaibal Barua}[%
	orcid=0000-0002-7305-7169,
	email=shaibal.barua@mdu.se,
	url=http://www.es.mdh.se/staff/2754-Shaibal\_Barua
	]
	\author[1]{Shahina Begum}[%
	orcid=0000-0002-1212-7637,
	email=shahina.begum@mdu.se,
	url=http://www.es.mdh.se/staff/146-Shahina
	]
	\author[1]{Mir Riyanul Islam}[%
	orcid=0000-0003-0730-4405,
	email=mir.riyanul.islam@mdu.se,
	url=http://www.es.mdh.se/staff/3845-Mir\_Riyanul\_Islam
	]
	\author[2]{Rosina O Weber}[%
	orcid=0000-0001-7048-8812,
	email=rosina@drexel.edu,
	url=https://www.cs.drexel.edu/~rw37/
	]
	\cormark[1]
	%%\fnmark[1]
	\address[1]{Mälardalen University, Västerås, Sweden}
	
	\address[2]{Drexel University, Philadelphia, PA, 19802, USA}
	
	\cortext[1]{Corresponding author. Authors are listed in alphabetical order.}
	
	\begin{abstract}
		%%The abstract should briefly summarize the contents of paper in 15--250 words.
		AI methods referred to as interpretable are often discredited as inaccurate by supporters of the existence of a trade-off between interpretability and accuracy. In many problem contexts however this trade-off does not hold. This paper discusses a regression problem context to predict flight take-off delays where the most accurate data regression model was trained via the XGBoost implementation of gradient boosted decision trees. While building an XGB-CBR Twin and converting the XGBoost feature importance into global weights in the CBR model, the resultant CBR model alone provides the most accurate local prediction, maintains the global importance to provide a global explanation of the model, and offers the most interpretable representation for local explanations. This resultant CBR model becomes a benchmark of accuracy and interpretability for this problem context, and hence it is used to evaluate the two additive feature attribute methods SHAP and LIME to explain the XGBoost regression model. The results with respect to {\it local accuracy} and {\it feature attribution} lead to potentially valuable future work.
	\end{abstract}
	
	\maketitle
	
	% if there is room, enter the nDCG formula or a description of how it is computed
	%%%check all modified (included and removed items and comments from discussion and conclusions)
	%%%add future work
	\setlength{\parskip}{0pt plus1pt minus1pt}
	\section{Introduction}
	Case-based reasoning (CBR) is considered an interpretable model given its typical adoption of the weighted Euclidean Distance to implement k-nearest neighbors. With this approach, the weights are usually associated with global features, affording model interpretability. The concentration of the learning in global weights can however limit CBR accuracy, thus helping support the claim of the existence of a trade-off between accuracy and interpretability \cite{gunning2019xai}.
	
	In explainable artificial intelligence (XAI), the trade-off between accuracy and interpretability has been debunked in different problem contexts with different data types. For example, using image data from mammograms, Barnett et al. \cite{barnett2021case} learned about deficiencies in their classifier when told by experts the classification was being done for the wrong reasons. When aligning the interpretable features with domain knowledge, the resultant interpretable model was more accurate than before. The trade-off claim is even more often dismissed when data is tabular (e.g., \cite{liu2022fast}). Notwithstanding, as it often happens in science, this claim has motivated valuable works such as the ANN-CBR Twins \cite{kenny2021explaining} where an accurate artificial neural network (ANN) is twinned with CBR as a presumed less accurate but interpretable model. The successful demonstrations of ANN-CBR Twins (ibid.) make this a valuable approach for exemplar-based explainability.
	
	This paper investigates the problem context of predicting flight delays. Air Traffic Flow Management (ATFM) costs, on average, approximately 100 Euros per minute for airlines \cite{cook2011european}. According to the FAA report in 2019\footnote{\url{https://www.faa.gov/data_research/aviation_data_statistics/media/cost_delay_estimates.pdf}}, the estimated cost due to delay, considering airlines, passengers, lost demand, and indirect costs, was thirty-three billion dollars. This high cost justifies the increased interest in predicting take-off time and delays \cite{dalmau2021explainable}.
	
	The take-off time is one of the root indicators of the delay of an aircraft as it propagates to all transportation networks, hence predicting it is key to enhancing air traffic. Predicting the delay of take-off time is a regression problem, where feature sets (both numeric and categorical) are used from flight plans, weather reports, and airline information. Departure delay has been characterized considering the spatial and temporal aspects ({\it e.g}., \cite{rebollo2014characterization,kim2016deep,yu2019flight,tran2020taxi,kovarik2020comparative,dalmau2019improving}). The methods used for predicting tasks in ATFM include neural networks (NN), random forest, gradient boosting machines, support vector machines, and linear regression \cite{kovarik2020comparative}.
	
	This paper describes a study whose starting point was to use flight data to predict departure delays using XGBoost via regression. XGBoost \cite{chen2016xgboost} is an implementation of gradient boosted decision trees (GBDT), an ensemble method that uses gradients to build highly accurate decision trees. This ensemble aspect limits the local interpretability of GBDT but still produces global importance factors that can make the model globally interpretable. For local interpretability, an alternative would be to adapt the ANN-CBR twins approach into a XGB-CBR. One of the twins steps is to extract from the non-interpretable (and presumably more accurate) method the representation that supports its accuracy and transfer it over to CBR. The XGBoost importance factors facilitate this step. However, when doing this, as detailed later, the CBR model alone using XGBoost importance factors as global weights, produced a smaller mean absolute error (MAE) than the original XGBoost regression model.
	
	The CBR model is more accurate ({\it i.e}., lower MAE), offers global interpretability, and interpretable local explanations. This justifies its use as a benchmark against which to evaluate explanation methods for XGBoost. We adopt two additive feature attribute methods, namely, SHAP \cite{lundberg2017unified} and LIME \cite{ribeiro2016should} to produce features to explain the XGBoost regression model.  
	
	One of the benefits of having CBR as the most accurate model is interpretability. Another benefit stems from the use of global weights for each feature. One important aspect when predicting air traffic delays is that some features are clearly more important than others, making the opportunity to incorporate domain knowledge desirable. For example, the feature that represents delays on the previous leg of a flight that uses the same aircraft is certainly relevant. Having only one weight for each feature makes it easy to incorporate or manipulate this kind of domain knowledge by directly changing the weight value. 
	
	%Section 2 introduces XGBoost \cite{chen2016xgboost} and CBR, followed by SHAP \cite{lundberg2017unified} and LIME \cite{ribeiro2016should}. Section 3 describes this paper methodology to build the data and explanation models, while Section 4 presents results and discussion, and Section 5 concludes.
	Section 2 introduces the methods and Section 3 describes this paper's methodology. Section 4 presents results and discussion, and Section 5 concludes.
	
	\section{Data and Explanation Models}
	This section describes the models discussed in this paper. The context is a regression model $r(x_{i})$ that uses data where $x \in X$ are instances mapped by features $f_j \in F$, $f_j = 1,\ldots, m$, $X_{\rm train} \subset X$ are training instances $x_i$, $x_i = 1,\ldots, n$ that include prediction delays $y \in Y$ in minutes, which are used by the regression model $r(x_{i})$ to learn predictions $\hat{y_i}$. $X_{\rm test} \subset X$ are testing instances.
	
	\subsection{Regression Models}
	
	XGBoost \cite{chen2016xgboost} is a GBDT ensemble method. Ensemble methods are shown to produce better performance than single methods \cite{sagi2018ensemble}. GBDT is an ensemble method for decision trees that learns with differentiable loss functions \cite{zhang2020gbdt}. Two GBDT variants are XGBoost \cite{chen2016xgboost} and LightGBM \cite{ke2017lightgbm}. XGBoost uses the second-order gradient to improve accuracy whereas LightGBM aims at improved efficiency. Previous work in air traffic delay prediction has utilized LightGBM \cite{dalmau2021explainable}. Hence, we start with XGBoost given its potential to be more accurate than LightGBM.
	
	CBR \cite{richter2013case} has its roots in memory-based methods from cognitive science \cite{schank1983dynamic}. CBR implements the similarity heuristic, \textit{i.e}., to reuse a previous solution to solve a similar new problem. Determining similarity between problems is domain-dependent, hence CBR systems often use the weighted Euclidean Distance where weights can reflect particular aspects of the problem context. 
	These weights used in similarity assessment are global to features, making decisions interpretable at the global level \footnote{Authors note that it is not within the scope of this paper to debate about the value of local versus global interpretability, but simply to point out when discussed interpretability is local or global.}. The limitation is that only global weights may limit accuracy. On the other hand, this simple and global representation facilitates incorporation of domain knowledge.
	When using the weighted Euclidean Distance, weights can be learned in various ways such as feedback learning algorithms \cite{Aha1998} or decision trees ({\it e.g.}, \cite{Gunaydin}). In this paper, the CBR model uses the XGBoost feature importance values as weights. 
	
	%One advantage of CBR is that it can be implemented with different similarity algorithms including neural networks, which will allow the use of more weights and then limit %its interpretability. It has been recently shown that the similarity heuristic from CBR combined with the notion that cases can be composed of prototypical features can be %successfully incorporated into interpretable deep learning architectures \cite{li2018deep,chen2019looks,barnett2021case}.
	
	\subsection{Explanation Methods}
	
	ANN-CBR Twins is an example-based explanation method \cite{kenny2019twin,kenny2021explaining}. The concept of Twins is based on the premise of two models where the accuracy-interpretability trade-off holds. The black-box and highly accurate ANN is one twin and the other is CBR, as the interpretable and less accurate model. The goal is that the models are functionally equivalent, that is, that they can produce the same results for the same testing instances. ANN-CBR twins succeed by transferring the representation and weights from the ANN into CBR \cite{kenny2019twin}.
	
	\subsubsection{Additive Feature Attribution}\label{afa}
	Explanation methods based on approaches to distribute gain in coalitional game theory \cite{strumbelj2010efficient,lundberg2017unified} utilize Shapley values \cite{shapley1997value} thus inheriting their properties. Lundberg and Lee \cite{lundberg2017unified} identify a class of explanation methods called \textit{additive feature attribution}, which include those based on Shapley values, among others \cite{lundberg2017unified}. This class is referred to as additive because of the efficiency property from Shapley values \cite{shapley1997value} that shows that the gains shared by all players in a coalition game equals the value of the grand coalition. This property becomes {\it local accuracy} for additive feature attribute methods (Equation 1) where $g(\boldsymbol{z_j})$ is the explanation model where the property of {\it local accuracy} is demonstrated when $g(\boldsymbol{z_j})$ matches the model $r(x_{i})$ for each instance, where $g(\boldsymbol{z_j})$ is computed on the vector $\boldsymbol{z_j}$ which transforms $\boldsymbol{x_j}$ by the function $h(\boldsymbol{z_j})$ makings $\boldsymbol{z_j} \in \{0,1\}^m$:
	\begin{equation}
		\label{e:g-model}
		g(\boldsymbol{z_j})  = \phi_0 + \sum_{j=1}^{j=m} \phi_{j} \boldsymbol{z_j}
	\end{equation}
	
	The local interpretable model-agnostic explanation (LIME) \cite{ribeiro2016should} is another additive feature attribution method. LIME fits a linear regression to explain the behavior of a sample point. To obtain  points for fitting a linear regression, LIME randomly perturbs the point to be explained using the points closest to the target point. The coefficients of the linear regression in LIME are used to produce $\phi$ values for Equation \ref{e:g-model} and predict the output of the model $g(\boldsymbol{z_j})$. 
	
	\subsubsection{XAI for Regression}
	Letzgus et al. \cite{letzgus2022toward} examine  XAI methods for regression problems. They recommend that both prediction and explanation be done with methods that do not normalize their values in order to preserve the alignment between the sum of the contributions with the prediction thus preserving the same measurement unit. They refer to it as the {\it conservation principle}.
	
	\section{Methodology}
	This section analyzes SHAP and LIME in terms of {\it local accuracy} and feature attribution for the XGBoost implementation for predicting flight delays. XGBoost predictions are the baseline for {\it local accuracy} because the explanation models were built for it; CBR is the baseline for feature importance because it is the most accurate model and it allows local interpretability.
	
	\subsection{Data} The dataset was collected and processed by EUROCONTROL\footnote{\url{https://www.eurocontrol.int/}} and it uses the Enhanced Tactical Flow Management System (ETFMS) flight data messages for all flights during the year 2019 ({\it i.e.}, May to October). The datasets include basic information, status of the flight and previous flight leg, ATFM regulations, weather, and calendar. The features are described in detail in \cite{dalmau2021explainable}.
	
	The data used for XGBoost includes 5,903,743 instances of the clean dataset with months from May to August, which is a subset of the dataset from EUROCONTROL. The study includes the first five days of September and October for testing, without using the remaining days of these months. The number of instances in the testing data is 158,147. The main difference between the data used in this paper and in \cite{dalmau2021explainable} is that they broke down the data into eight intervals of time to EOBT (Estimated Off-Block Time). In this paper, the data was not broken down in intervals, which means using the interval from zero to three-hundred and sixty minutes: (0,360].
	
	\subsection{Metrics}We use MAE and standard deviation $\sigma$ for the quality of the predictions for both data and explanation models. MAE computes the average difference between an actual observation and a prediction from a model:
	\begin{equation}
		\label{e:mae}
		MAE = 1/n  \sum_{i=1}^{i=n}  \lvert y_i - \hat{y_i} \rvert 
	\end{equation}
	
	\paragraph{MAE for Data Models.} MAE is computed based on the actual delays $y_i$ from the testing data as baseline for comparison against the predictions $\hat{y_i}$ learned by the regression models $r(x_{i})$.
	
	\paragraph{MAE for Explanation Models.} As described in Section \ref{afa}, both SHAP and LIME use a function  $g(\boldsymbol{z_j})$ to produce a prediction $\hat{y}'
	\in Y$ using Equation \ref{e:g-model}. The values for MAE for the two explanation models are obtained from the difference between the predictions $\hat{y_i}$ learned by the regression model $r(x_{i})$ and the $\hat{y}'$ obtained by $g(\boldsymbol{z_j})$.
	
	%\paragraph{Normalized Discounted Cumulative Gain(nDCG)}nDCG is a variant of the DCG metric designed to compare the order of retrieved documents in information retrieval. Studies \cite{busa2012apple,wang2013theoretical} have demonstrated that it is possible to obtain different results when using nDCG. In this paper, we computed nDCG with sklearn library \cite{scikit-learn}. 
	
	\paragraph{Normalized Discounted Cumulative Gain (nDCG).} nDCG compares the order of retrieved documents in information retrieval. Studies \cite{busa2012apple,wang2013theoretical} show that different libraries can produce varied results. In this paper, we computed nDCG with sklearn library \cite{scikit-learn}. 
	
	\subsection{Methods}
	\paragraph{XGBoost}The hyperparameters for XGBoost were selected based on the results from 288 different combinations. The final model used the following: {learning\_rate} = 0.1, {max\_depth} = 7, {min\_child\_weight} = 1, {subsample} = 0.5, {colsample\_bytree} = 0.5, {n\_estimators} = 500.
	
	\paragraph{CBR}The CBR model averages the predictions in the three least distant neighbors retrieved using the Euclidean Distance weighted with the XGBoost importance factors. For binary and categorical features, local similarity is symbolic producing $1$ when values are equal and $0$ when different. For numeric features, the absolute difference is divided by the range of values. As a local learner, the predictions are computed with leave-one-out cross validation. 
	%%consider making the formula for the euclidean distance expression with weights
	
	\paragraph{Additive and Global CBR} CBR can be used for example-based explanations, but its global weights do not support local explanations in the same form as additive models. CBR global weights support global interpretability, which we refer as Global CBR. Additive CBR is an additive version built by re-scaling the values for the CBR regression model after prediction. Additive CBR becomes a benchmark for local interpretability. Feature values and weights are re-scaled to produce $\phi_j\boldsymbol{z_j}^{CBR}$ in the same terms as the additive feature attribution explanation models $g(\boldsymbol{z_j})$. To achieve this, we utilize a multiplier $\gamma_i$ obtained by dividing the prediction $\hat{y_i}$ of the CBR regression model $r(x_{i})$ by the sum of its factors $x_{ij}w_j$ using Equation \ref{e:addcbr}\footnote{The previous version contained an erroneous sum in Equation \ref{e:addcbr} that we have corrected in this version.}:
	
	\begin{equation}
		\label{e:addcbr}
		\gamma_i = \hat{y_i} 1/(\sum_{j=1}^{m} x_{ij} w_j) 
	\end{equation}
	
	\paragraph{SHAP and LIME}The two explanation models were built for XGBoost. SHAP was implemented using kernelSHAP with default settings. LIME was implemented with 1,000 perturbations and 1,000 number of samples. 
	
	\section{Results and Discussion}\label{Sec4}
	\subsection{Results from Data Models}
	
	\begin{table*}[h]
		\caption{Average MAE and standard deviation $\sigma$ for data models on three sets of the data. {\it All} uses the 158,147 instances, {\it 100k} and {\it 67k} use, respectively, the 100,650 and 67,495 most accurate testing instances.}
		\label{tab:MAEdatamodels}
		\begin{tabular}{c|ccc|ccc}
			\toprule
			Model & \multicolumn{3}{c|}{CBR}  & \multicolumn{3}{c}{XGB} \\
			Instances & All& 100k& 67k& All& 100k& 67k\\
			\midrule
			Average MAE   &5.88  & 1.48 & 0.52 & 9.22 &4.28& 2.72  \\
			$\sigma$ MAE   &8.22  & 1.55 & 0.70 & 8.91 &2.67& 1.62  \\
			\bottomrule
		\end{tabular}
	\end{table*}
	
	The first results in this section demonstrate the basis for using CBR as the baseline and justify why the XGB-CBR Twin is not required.
	The results for MAE and standard deviation for the XGBoost and CBR data models are shown in Table \ref{tab:MAEdatamodels}.
	
	\begin{figure}[h]
		\centering
		\includegraphics[height=4cm,keepaspectratio]{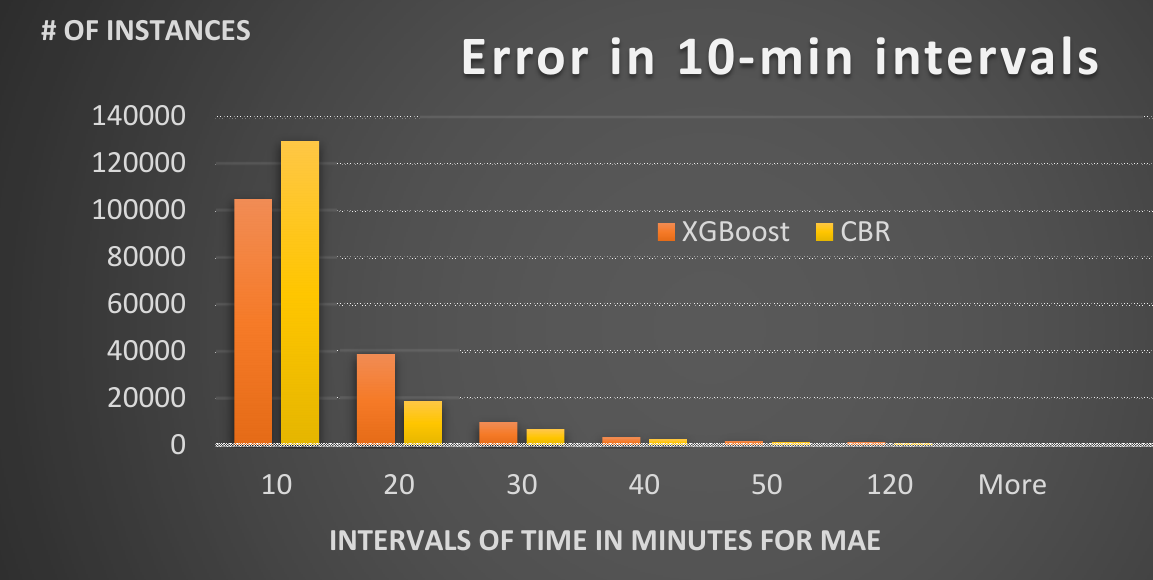}
		\caption{Histograms grouping the number of instances within bins of MAE for CBR and XGBoost}
		\label{tab:histcbr}
	\end{figure}
	
	\begin{figure}[h]
		\centering
		\includegraphics[height=4cm,keepaspectratio]{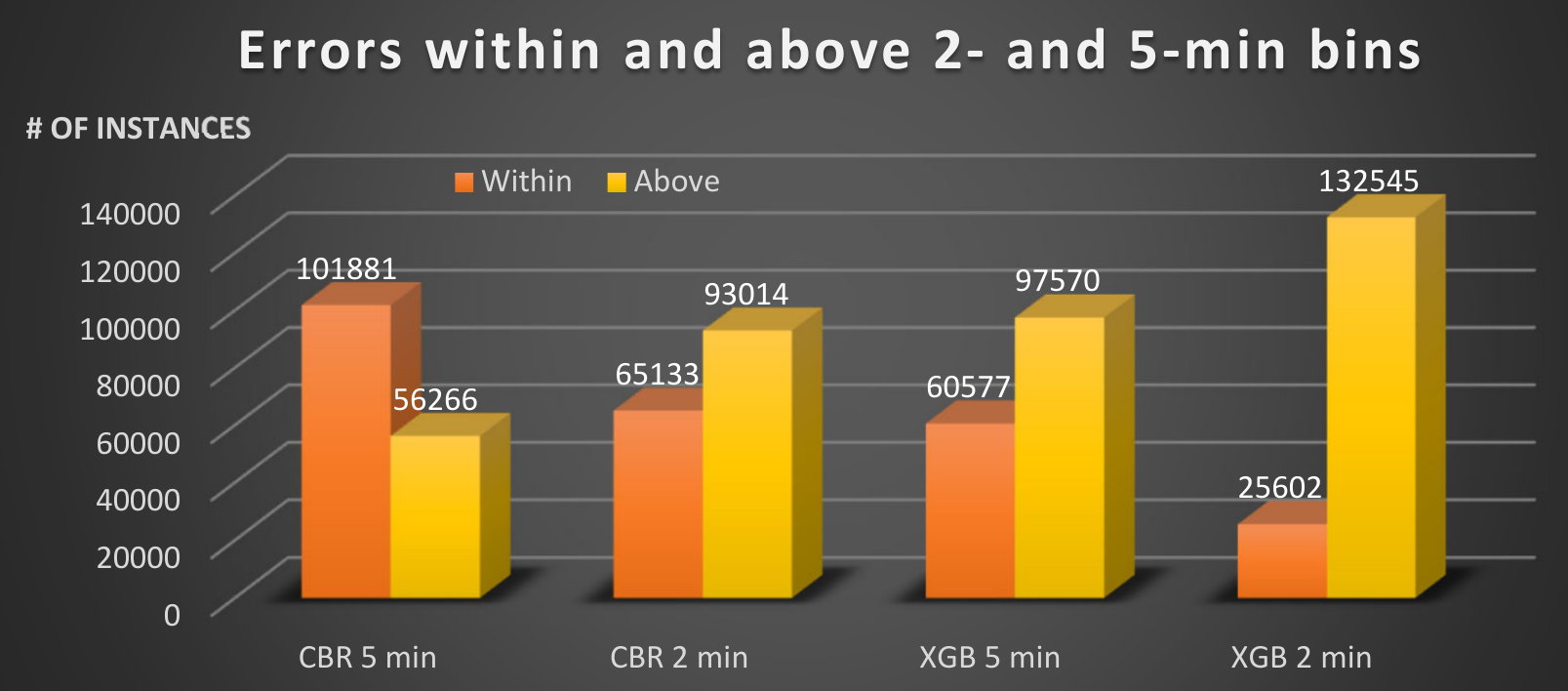}
		\caption{Instances with errors within and above 2 and 5 minutes}
		\label{tab:25MIN}
	\end{figure}
	
	Because CBR performance challenges the accuracy versus interpretability trade-off \cite{gunning2019xai}, this section provides details for the average results from Table \ref{tab:MAEdatamodels}. The first results present how far the regression predictions are from the actual delays. The histograms in Figure \ref{tab:histcbr} show that the lower error in the CBR model is based on the model having more instances with lower errors. These lower errors were within the 10-minute range. The XGBoost model has fewer errors in the bin of 10 minutes and more in the bins with higher errors, leading to greater values in MAE.
	
	Figure \ref{tab:25MIN} depicts the number of instances at the five- and two-minute marks for CBR and XGBoost. At these thresholds, the CBR model produces more instances within five minutes difference from the actual predictions than with higher errors. At the two-minute mark, CBR has about 40\% of instances within two minutes away from the actual prediction.
	
	\subsection{Discussion on Data Models}
	The higher CBR accuracy incites the question as to whether CBR models would consistently benefit from learning global weights via ensemble models. These results allow the use of CBR as a baseline for explanation quality because it is both the most interpretable and most accurate. This would represent one circumstance in which it would not be necessary to adopt the Twins approach. Had the CBR model not been the most accurate, using Twins would be preferable. 
	
	%consider remving the first part of this following paragraphif we need space
	%A comparison between the results presented in this paper and those from Dalmau et al. \cite{dalmau2021explainable} would not be valid given Dalmau et al. predicted delays within different time intervals and the study here only predicted delays for all the data. The smallest error in Dalmau et al. was for the shortest interval (0, 15], with a MAE of 8.8 and standard deviation of 19.6 minutes using LightGBM. The MAE obtained with CBR was 5.88 minutes with 8.22 standard deviation, both lower when using only the test data. These results suggest multiple considerations such as the need to create multiple variations of the test data and obtain the average and standard deviations of all those datasets to confirm the results in this paper are not chance. It would be useful to make predictions using the same intervals as used in \cite{dalmau2021explainable}, as well as experiment with other methods not discussed herein. 
	
	\subsection{Results from Explanation Models}
	
	\begin{table*}[h]
		\caption{Average MAE and standard deviation $\sigma$ for {\it local accuracy} of explanation models on three sets of instances.}
		\label{tab:localacc}
		\begin{tabular}{c|cc|cc|cc}
			\toprule
			\# of instances &  \multicolumn{2}{c|}{ALL}  & \multicolumn{2}{c|}{100k}  &  \multicolumn{2}{c}{67k}\\
			Model & SHAP & LIME & SHAP & LIME & SHAP & LIME  \\
			\midrule
			average MAE & \num{3.3e-6} & 8.62& \num{1.1e-6} & 4.75 & \num{6.2e-7} & 3.13  \\
			$\sigma$ MAE & \num{4.3e-6} & 6.32& \num{7.9e-7} & 2.81 & \num{4.0e-7} & 1.82  \\
			\bottomrule
		\end{tabular}
	\end{table*}
	
	\begin{table*}[h]
		\caption{Pairwise comparison of nDCG for SHAP and LIME across the three sets against the two baselines Global and Additive CBR for all 42 features. Higher is better. Highest value is in bold.}
		\label{tab:ndcg}
		\begin{tabular}{c|ccc|ccc}
			\toprule
			Baseline & \multicolumn{3}{c|}{Global CBR}  &   \multicolumn{3}{c}{Additive CBR}\\
			\# of instances & All & 100k & 67k & All & 100k & 67k   \\
			\midrule
			SHAP & 0.74 & 0.72 & 0.72 & 0.81 & \textbf{0.81} & \textbf{0.82}  \\
			LIME & \textbf{0.88} & \textbf{0.85} & \textbf{0.80} & \textbf{0.82} & 0.81 & 0.77  \\
			\bottomrule
		\end{tabular}
	\end{table*}
	
	Table \ref{tab:localacc} includes average and standard deviation MAE of  {\it local accuracy} for SHAP and LIME with respect to XGBoost, the model for which the local explanations were built. Table \ref{tab:ndcg} presents the nDCG values comparing the order of feature attributions. Table \ref{tab:progress} combines the two previous tables to show the progression of values. We observe that Tables \ref{tab:ndcg} and \ref{tab:progress} include comparisons against Global CBR for reference purposes, but the intended benchmark for analysis is Additive CBR because it is formulated as an additive model.
	
	\subsection{Discussion on Explanation Models}
	Table \ref{tab:localacc} shows the impressive {\it local accuracy} obtained by SHAP. These MAE values correspond to precision levels of $10^{-5}$ and $10^{-6}$. LIME produces average error still above three minutes in the smallest set of instances with lowest MAE. This difference is not observed in the analysis of feature attributions in Table \ref{tab:ndcg}. This difference may be explained by a few aspects.
	
	At the smallest set with the most accurate instances (Table \ref{tab:ndcg}), SHAP's attributions provide higher nDCG values ({\it i.e}., 0.82) than LIME ({\it i.e}., 0.77). This result is not as impressive as results for {\it local accuracy} but shows SHAP as superior. The fact that SHAP does not have higher nDCG values may be because local explanations are built to model the data model, which is XGBoost, not CBR. As it can be seen in Table \ref{tab:MAEdatamodels}, there is reasonable difference between the MAE of CBR ({\it i.e}., 0.52) and XGBoost ({\it i.e}., 2.72) with respect to the actual data at the set of instances with lowest MAE. This variation might explain why the nDCG values for SHAP are not higher.

	\begin{table*}[h]
		\caption{Progression of both {\it local accuracy} in MAE and nDCG for SHAP and LIME. nDCG(Gl) is compared against Global CBR, nDCG(Add) is compared against Additive CBR. Rows show the number of instances. For {\it local accuracy} in MAE, lower is better. For nDCG, higher is better. Best value is in bold.}
		\label{tab:progress}
		\begin{tabular}{ccccccc}
			\toprule
			XAI Model & \multicolumn{3}{c}{SHAP}  &   \multicolumn{3}{c}{LIME}\\
			Metric & {\it local accuracy} & nDCG(Add) & nDCG(Gl) & {\it local accuracy} & nDCG(Add) & nDCG(Gl)\\
			\midrule
			All & \num{3.3e-6} & 0.806 & \textbf{0.806} & 8.62 & \textbf{0.819} & \textbf{0.882}  \\
			100k & \num{1.1e-6} & 0.813 & 0.722 & 4.75 & 0.805 & 0.847 \\
			67k & \textbf{\num{6.2e-7}} & \textbf{0.817} & 0.717 & \textbf{3.13} & 0.773 & 0.800  \\
			\bottomrule
		\end{tabular}
	\end{table*}
	
	Table \ref{tab:progress} includes the values for {\it local accuracy} for easy examination of their progression. Moving from the data set with all instances, which is expected to be the least accurate, SHAP's {\it local accuracy} improves going from the first ({\it i.e}., 0.806), to second ({\it i.e}., 0.813), and third row ({\it i.e}., 0.817), showing {\it local accuracy} and nDCG are somehow proportional. nDCG values for LIME are inversely proportional, decreasing from the first ({\it i.e}., 0.819), to second ({\it i.e}., 0.805), and third row ({\it i.e}., 0.773). One possible observation is that LIME's low {\it local accuracy} is consistent with lack of progression of nDCG. In any case, these results suggest further studies are needed because they do not provide the means to support that any of these feature attributions is valid. We list three aspects to investigate: feature attribution, {\it local accuracy}, and additive variants.
	
	\paragraph{Feature Importance and Feature Attribution}The literature indicates that the contribution of a feature in an additive feature attribution model is different from feature importance in the sense of weights \cite{letzgus2022toward}. The question arises on whether there are any relations to be drawn between these two types of feature importance. One direction would be to question whether example-based explanations produced by CBR support feature attributions resultant from any explanation model. Another would be on whether there is any relationship between feature importance in the sense of weights as practiced in CBR and feature attributions based on contributions of an additive model. Further studies are needed to shed light into the claim that the “\textit{best explanation of a simple model is the model itself}” \cite{lundberg2017unified} pp 2. If the explanation method models the data model decision boundary then what information content does it produce? If the explanation method models instance points, then what does it mean for a feature to contribute to a decision? Questions such as how to precisely define feature attributions and feature importance are crucial to support proper presentation of XAI results to users.
	
	\paragraph{Local Accuracy}The recommendation is to investigate whether {\it local accuracy} is an indicator of feature attribution quality. This study, of course, depends on a definition of feature attribution.
	
	\paragraph{Additive Variants}It is not clear if the Additive CBR model adopted herein meets the {\it conservation principle} \cite{letzgus2022toward} for regression and is thus valid as a benchmark. A review of the literature should clarify which data models can be re-scaled into additive models to enable valid comparisons.
	
	\section{Related Works}
	Many papers have attempted to evaluate and comparatively analyze explanation methods (\textit{e.g.}, \cite{adebayo2018sanity,man2020best,zhou2021feature}). There are multiple ways to categorize explainable methods, but a valuable, and often dismissed, perspective is to consider the information type an XAI method produces. Methods that reply to the question, “Why not something else?” produce counterfactual instances and cannot be included in the same category as feature attribution methods, which aim to produce contributions of instance features. This paper compares two XAI methods that belong to the category of \textit{additive feature attribution methods} \cite{lundberg2017unified}, namely, SHAP and LIME. 
	
	Zhou et al. \cite{zhou2021feature} point out the fact that \textit{attribution} is not a well defined term as they compare additive (\textit{e.g.}, SHAP \cite{lundberg2017unified})  against non-additive methods such as (\textit{e.g.}, \cite{DBLP:journals/corr/SelvarajuDVCPB16,DBLP:journals/corr/SmilkovTKVW17}). Their rationale for the selection is that all these methods can be used to produce visualizations known as saliency maps. Zhou et al. \cite{zhou2021feature} propose to transform datasets as a means to create ground-truth data and assess whether these methods can succeed in recovering them. The authors conclude none of the methods can be considered satisfactory.
	
	The benefit of limiting the set of methods to evaluate lies on the ability to compare along the same deliverable. \textit{Additive feature attribution methods} \cite{lundberg2017unified} share the same properties and thus using the features they identify with highest importance and their {\it local accuracy} seem a reasonable starting point. As recommended by various authors  (\textit{e.g.}, \cite{sundararajan2017axiomatic,yang2019benchmarking,deyoung2019eraser,amiri2020data}) the use of benchmark datasets is valid as long as the evaluation is limited to feature importance or {\it local accuracy}. As previously described \cite{yang2019evaluating}, benchmark datasets are not recommended for evaluating explanations for user consumption because explanations are user-, context-, and application-specific ({\it e.g.}, \cite{gunning2019xai,arrieta2020explainable,mueller2019explanation}).
	
	\section{Concluding Remarks and Future Works}
	This paper describes a regression problem for air traffic delay prediction where an interpretable data model is also the most accurate, hence demonstrating another instance where the accuracy-interpretability trade-off does not hold. Here the study built a reasonably accurate model (\textit{i.e.}, MAE 9.22) with XGBoost and wanted to have a more interpretable model by building an XGB-CBR twin. When transferring the importance factors from XGBoost into CBR as global weights, the CBR model turned out to be even more accurate (\textit{i.e.}, MAE 5.82) than the XGBoost. The study then used the interpretable CBR model as a benchmark to compare the performance of the two additive feature attribution methods SHAP and LIME. The selection of these two methods was based on their {\it local accuracy} property where each explanation model is able to produce a prediction just like the regression model. When examining {\it local accuracy}, the SHAP explanation model was functionally equivalent to the original XGBoost model, predicting the same delays at a precision of $10^{-5}$. The MAE between LIME and XGBoost is 8.62 minutes.
	
	Based on the assertion that the best explanation of a model is the model itself \cite{lundberg2017unified}, the results compare whether the level of equivalence between the data and the explanation models could translate into feature attribution quality. Nonetheless, when comparing the feature importance from the Additive CBR baseline against feature attributions from LIME and SHAP, SHAP's superior performance in {\it local accuracy} is not matched. Based on these results, a few questions arise with potential to advance the field (Section \ref{Sec4}).
	
	%future work {\it local accuracy}
	%cbr
	%The initial idea is to transfer the weights computed by XGBoost to CBR. This way CBR is as good as XGBoost. Here, the first question is about the roots of this "good" performance. Is it because XGBoost is not as good as other approaches (i.e. ANN)? Could another CBR-optimization technique (such as genetic algorithms) find the weights that make CBR better than XGBoost? And, more importantly, could you prove that this method (CBR-optimization with XGBoost weights) will also work (CBR achieves a similar accuracy) in other domains/datasets?
	
	Among important future work are studies on learning feature weights for CBR and identifying when Twins are preferable. For predicting flight delays, future work includes comparisons against the results from \cite{dalmau2021explainable}, other data models, and other additive feature attribution methods.

	\begin{acknowledgments}
		This work was financed by the project xApp: Explainable AI for Industrial Applications which is funded by VINNOVA (Sweden’s innovation agency) and Mälardalen University. The authors would like to acknowledge VINNOVA to support our research in the area of explainable AI (XAI) in the project “Application of Explainable AI in Industrial Applications (Diary number 2021-03971)”.
		This study is also a part of the project Transparent Artificial Intelligence and Automation to Air Traffic Management Systems, ARTIMATION, funded by the European Union’s Horizon 2020 within the framework SESAR 2020 research and innovation program under Grant Agreement N. 894238. The authors would like to thank Eurocontrol for providing the dataset for this study, and the reviewers for their excellent suggestions.
		
	\end{acknowledgments}
	
	%%
	%% Define the bibliography file to be used
	\bibliography{sample-ceur}
\end{document}